\documentclass{article}
\usepackage{amsmath,epsfig}
\usepackage[preprint]{spconfa4}

\usepackage{amsfonts}
\usepackage{booktabs}
\usepackage{multirow}
\usepackage{array}
\usepackage{caption}

\copyrightnotice{978-1-6654-3864-3/21/\$31.00 \copyright2021 IEEE}

\let\OLDthebibliography\thebibliography
\renewcommand\thebibliography[1]{
  \OLDthebibliography{#1}
  \setlength{\parskip}{0pt}
  \setlength{\itemsep}{0pt plus 0.3ex}
}

\begin{document}\sloppy

\def\x{{\mathbf x}}
\def\L{{\cal L}}
\def\etal{{\em et al.}}
\def\ie{{\em i.e. }}
\def\eg{{\em e.g. }}
\makeatother

\title{ LI-Net: Large-Pose Identity-Preserving Face Reenactment Network}

\name{Jin Liu$^{\ast,\dagger}$\thanks{This research is supported in part by  the National Key Research and Development Program of China under Grant No.2020AAA0140000.},Peng Chen$^{\ast,\dagger}$,Tao Liang$^{\ast,\dagger}$,Zhaoxing Li$^{\ast}$,Cai Yu$^{\ast,\dagger}$,Shuqiao Zou$^{\ast,\dagger}$,Jiao Dai$^{\ast}$,Jizhong Han$^{\ast}$}
\address{$^{\ast}$Institute of Information Engineering, Chinese Academy of Sciences, Beijing, China\\
	$^{\dagger}$School of Cyber Security, University of Chinese Academy of Sciences, Beijing, China\\
\{liujin, chenpeng, liangtao0305, lizhaoxing, caiyu, zoushuqiao, daijiao, hanjizhong\}@iie.ac.cn}

\maketitle

\begin{abstract}
	Face reenactment is a challenging task, as it is difficult to maintain accurate expression, pose and identity simultaneously. Most existing methods directly apply driving facial landmarks to reenact source faces and ignore the intrinsic gap between two identities, resulting in the identity mismatch issue. Besides, they neglect the entanglement of expression and pose features when encoding driving faces, leading to inaccurate expressions and visual artifacts on large-pose reenacted faces. To address these problems, we propose a Large-pose Identity-preserving face reenactment network, LI-Net. Specifically, the Landmark Transformer is adopted to adjust driving landmark images, which aims to narrow the identity gap between driving and source landmark images. Then the Face Rotation Module and the Expression Enhancing Generator decouple the transformed landmark image into pose and expression features, and reenact those attributes separately to generate identity-preserving faces with accurate expressions and poses. Both qualitative and quantitative experimental results demonstrate the superiority of our method.
\end{abstract}

\begin{keywords} 
	Face Reenactment, Image Synthesis, Generative Adversarial Network
	\vspace{-2mm}
\end{keywords}

\section{Introduction}
\vspace{-2mm}
\label{sec:intro}
Given a source face and driving face, \emph{face reenactment} aims to generate a reenacted face which is animated by the expression and pose of the driving face while preserving the identity of the source face. Various applications benefit from face reenactment, including telepresence, gaming and film-making. Recently, diverse face reenactment methods have emerged. However, existing methods suffer from identity mismatch and inaccurate expressions in large pose. In this work, we propose a Large-pose Identity-preserving face reenactment framework LI-Net to address the above problems.

Current face reenactment methods mainly rely on 3D models and GAN models. 3D-based methods reconstruct source faces using pre-defined 3DMM~\cite{blanz1999morphable} and render reenacted faces on the image plane, which suffers from tedious big-budget model construction procedure and high computation cost. For GAN-based methods, various approaches~\cite{wu2018reenactgan, zhu2017unpaired} apply simple encoder-decoder network to reenact faces, restricted by predefined identities. Later, diverse many-to-many methods arise, which reenact any \emph{unseen} identity using motion information from \emph{unseen} driving faces. They either utilize large-scale data to improve generalization~\cite{siarohin2019first} or adopt image-warping to make full use of facial details in source faces ~\cite{wiles2018x2face}. Nonetheless, the above methods ignore the intrinsic gap between two identities and directly apply driving faces to reenact source faces, leading to identity mismatch problem. Besides, they reenact new expressions and poses simultaneously with the entanglement of these attributes, leading to inaccurate expressions and visual artifacts in large pose.

\begin{figure}[t] 
	\includegraphics[width=\linewidth]{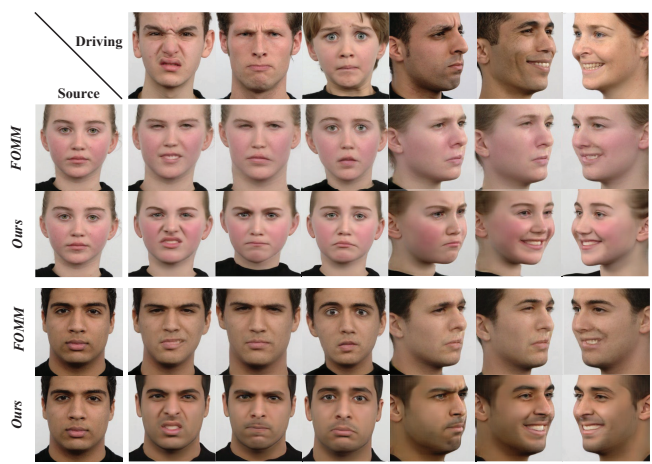} 
	\caption{\textit{\textbf{The results of LI-Net and current SOTA method FOMM~\cite{siarohin2019first}.} The first column and row are source and driving images. The other four rows are reenacted faces using methods notated on the left. Note we do not have \emph{identity mismatch} or \emph{inaccurate expressions} in large pose. }}
	\label{fig:problem}
	\vskip -0.2in
\end{figure}

\begin{figure*}[t] 
	\includegraphics[width=\linewidth]{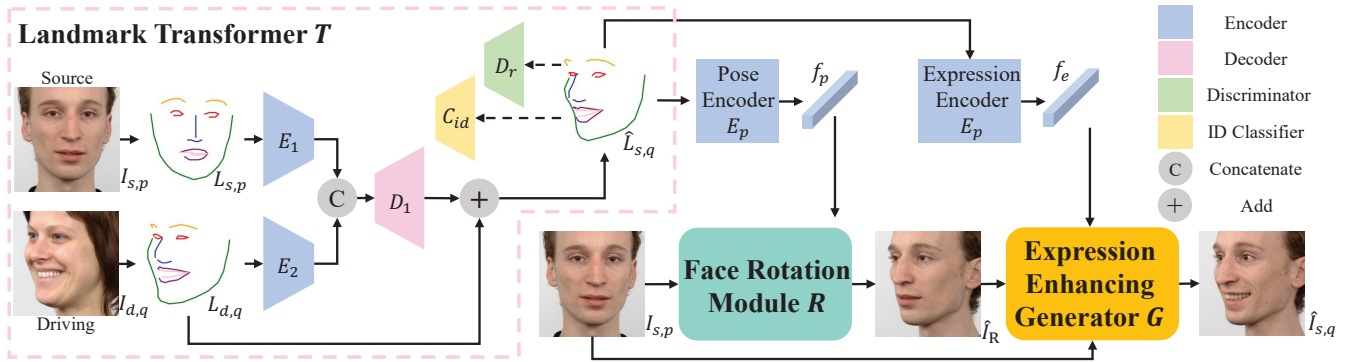} 
	\vspace{-7mm}
	\caption{\textit{\textbf{Overview of the proposed LI-Net.} First, the landmark transformer $ T $ (Section \ref{sec: transformer}) takes $L_{s,p}$ and  $L_{d,q}$ as inputs to generate transformed landmarks $\hat{L}_{s,q}$. The discriminator $ D_{r} $ and ID classifier $ C_{id} $ constrain the realness and identity of $\hat{L}_{s,q}$. Second, the face rotation module $ R $ (Section \ref{sec: rotation}) will rotate source images according to pose information from $\hat{L}_{s,q}$. Finally, the expression enhancing generator $ G $ (Section \ref{sec: enhancing}) will enhance $ \hat{I}_{R} $ and add expressions to get reenacted faces  $\hat{I}_{s,q}$.}}
	\label{fig:flowchart}
	\vspace{-3mm}
\end{figure*}

In detail, \emph{identity mismatch} means the inability of face reenactment model to preserve the identity of the source face. We take the SOTA method FOMM~\cite{siarohin2019first} as an example. As shown in the first three columns in Fig.~\ref{fig:problem}, we cannot tell the source identity from reenacted faces of FOMM whose facial contours are exactly the same as driving faces instead of source ones. Furthermore,  when it comes to large-pose reenactment, \emph{inaccurate expressions} and \emph{visual artifacts} appear. Fuzzy mouth shape and weird facial contour can be found in FOMM results, as shown in the last three columns of Fig.~\ref{fig:problem}.

To address the above problems, we propose a Large-pose Identity-preserving face reenactment network called LI-Net. Specifically, given source and driving landmark images, the \emph{Landmark Transformer} is utilized to transform driving landmarks to source identities.
The \emph{Face Rotation Module} then takes source face and transformed landmarks as input to generate rotated source faces with driving poses. Finally, source faces, rotated faces and transformed landmarks will be sent into \emph{Expression Enhancing Generator} to get reenacted faces.

To the best of our knowledge, the proposed LI-Net is the first to perform many-to-many identity-preserving face reenactment while manages to maintain accurate identity, pose and expression simultaneously. Our contributions are as follows: (a) To solve the identity mismatch problem, the landmark transformer with explicit identity restrictions is proposed to transform driving landmarks to source identities. (b) We adopt the face rotation module with a novel data augmentation method and apply the expression enhancing generator to impose fine-grained pose and expression control over reenacted faces. (c) Both qualitative and quantitative experimental results indicate that the proposed method manages to reenact large-pose identity-preserving high-quality faces with accurate expressions and photo-realistic facial details.
\vspace{-2mm}

\pagestyle{empty}  
\thispagestyle{empty} 

\section{Related work}
\vspace{-2mm}
\textbf{Conditional Generative Adversarial Network.} Diverse methods related to GAN~\cite{NIPS2014_5423} are proposed to achieve facial attribute editing or whole face image generation. cGAN~\cite{mirza2014conditional} controls the attribute of faces using condition information vectors. Pix2Pix~\cite{pix2pix2017} takes face sketch photo as conditions to generate faces. MaskGAN~\cite{CelebAMask-HQ} edits faces by mapping the segmentation mask to the target image. StyleGAN~\cite{karras2019style} applies transformation on latent codes to control face image styles using adaptive instance normalization (AdaIN)~\cite{huang2017arbitrary}. Unlike these methods, LI-Net maps input driving faces to landmark images, which owns more scalability than vectors and is easy to obtain compared with segmentation masks.

\noindent\textbf{Face Reenactment.} Face reenactment mainly falls in two categories, 3D-based and GAN-based methods. 3D-based methods~\cite{thies2016face2face, suwajanakorn2017synthesizing} utilizes pre-defined 3DMM~\cite{blanz1999morphable} to reconstruct shape, expression, texture and illumination of input faces and reenact faces by changing the corresponding parameters, which always reflects specific human races collected to construct 3DMM. For GAN-based methods, ReenactGAN~\cite{wu2018reenactgan} maps face pose information to a boundary latent space and reenact faces by a person-specific generator, restricted by the specific person identity. FReeNet~\cite{zhang2020freenet} pre-processes landmarks and uses generator to apply many-to-many face reenactment, which can not transfer poses of driving faces, only expressions. X2Face~\cite{wiles2018x2face} produces embedded faces from source faces and utilizes an encoder-decoder architecture. FOMM~\cite{siarohin2019first} predicts optical flow and occlusion masks to reenact faces using image warping. These methods suffer from identity mismatch and inaccurate expressions in large-pose faces. However, to solve the above problems, LI-Net uses a landmark transformer with explicit identity constraints and applies separate modules to impose fine-grained control over reenacted faces, regarding poses and expressions.
\vspace{-8mm}

\section{METHODOLOGY}
\vspace{-2mm}
\label{sec:methodology}
The procedure of LI-Net is shown in Fig. \ref{fig:flowchart}. The landmark detector~\cite{bulat2017far} first obtains landmark images $ L_{s,p} $ and $ L_{d,q} $  from faces $ I_{s,p} $ and $ I_{d,q} $. $ I $ and $ L $ denote faces and landmark images, respectively. The first subscript denotes identity while the second means motion information (expression and pose). Then the landmark transformer $ T $ transforms driving landmarks to source identities. Subsequently, based on a novel data augmentation method, the face rotation module $ R $ generates $ \hat{I}_{R} $, containing pose information from $ \hat{L}_{s,q} $. Finally, the expression enhancing generator $ G $ generates the final reenacted face $ \hat{I}_{s,q} $, sharing the same pose and expression as $ I_{d,q} $.

\begin{figure}[t] 
	\includegraphics[width=\linewidth]{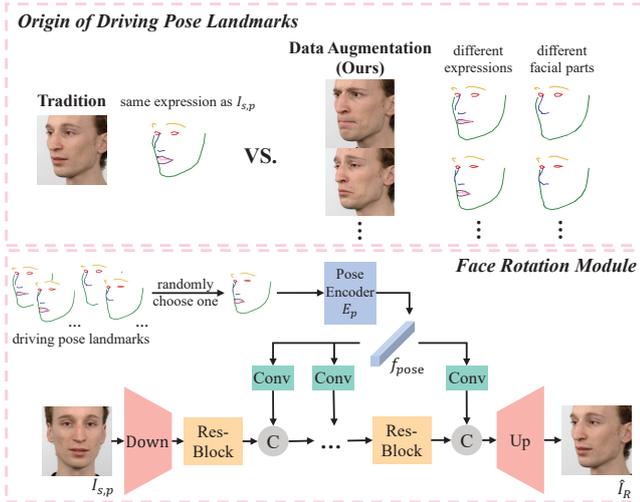} 
	\vspace{-7mm}
	\caption{\textit{\textbf{Face Rotation module $ R $ and driving pose landmark images used in $ R $.} For a source image $ I_{s,p} $ and its corresponding rotated image $ \hat{I}_{R} $ with driving pose, we extract pose features from face landmarks of same pose but with different expressions and facial parts. Each feature extracted will be sent into $ R $ to guide the same rotation process from $ I_{s,p} $ to $ \hat{I}_{R} $.}}
	\label{fig:rotator}
	\vspace{-3mm}
\end{figure}

\vspace{-2mm}
\subsection{Landmark Transformer}
\vspace{-1mm} 
\label{sec: transformer}
To solve the identity mismatch problem, we propose a landmark transformer to transform driving landmark images to source face identities. As shown in Fig. \ref{fig:flowchart}, source face landmark $L_{s,p}$ and driving face landmark $L_{d,q}$ are sent to corresponding encoder $E_{1}$ and $E_{2}$. Then the decoder $D_{1}$ transforms encoded features to predicted landmark shift, which will be added to $L_{d,q}$ and finally gets transformed reenacted landmarks $\hat{L}_{s,q}$. The landmark shift acts as fine adjustments on driving landmarks to give source identity information. Also, the identity classifier $C_{id}$ and discriminator $D_{r}$ guarantee the identity consistency and the realness of transformed landmarks. In general, we get the landmark transformer as $\hat{L}_{s,q} = T(L_{s,p}, L_{d,q})$, where the first and second input of $ T $ are source and driving landmarks, respectively. The overall loss function $\mathcal{L}_{T}$ is defined as:
\vspace{-2mm}
\begin{equation}
	\begin{aligned}
		\mathcal{L}_{T}=& \lambda_{L1}\mathcal{L}_{L1} + \lambda_{rec}\mathcal{L}_{rec} + \lambda_{cycle}\mathcal{L}_{cycle} +\\& \lambda_{id}\mathcal{L}_{id} +  \lambda_{D}\mathcal{L}_{D}, \label{equ:lt_total_loss}
	\end{aligned}
	\vspace{-2mm}
\end{equation}
where all $\lambda$ are weights of various loss terms. We define $\mathcal{L}_{L1}$ as pixel-wise L1 loss between transformed landmark $\hat{L}_{s,q}$ and the corresponding ground truth landmark ${L}_{s,q}$.

\noindent\textbf{Reconstruction Loss.} The landmark transformer should not only transform landmarks between two different identities, but also reconstruct the driving landmark when receiving different landmarks with same identity. which is constrained by:
\begin{equation}
	\mathcal{L}_{rec}=\left\|  T(L_{s,p}, L_{s,q}) - L_{s,q}   \right\|_{1}.
\end{equation}

\noindent\textbf{Cycle Loss.} Inspired by~\cite{zhu2017unpaired}, $ \mathcal{L}_{cycle} $ are added in order to guarantee that the transformed landmark $ \hat{L}_{s,q} $ can be transformed back again to improve the fidelity, as shown below:
\begin{equation}
	\mathcal{L}_{cycle}=\left\|  T( L_{d,p}, T(L_{s,p}, L_{d,q})) - L_{d,q}   \right\|_{1}.
\end{equation}

\noindent\textbf{Identity Loss.} We introduce an identity classifier $ C_{id} $ to extract id features and predict id labels. During training, we use the cross entropy loss between predicted and ground truth identity label to constrain the classifier. The identity loss term $ \mathcal{L}_{id} $ in Equation \ref{equ:lt_total_loss} is defined as the mean absolute error between identity features of $ L_{s,p} $ and $ \hat{L}_{s,q} $.

\noindent\textbf{Adversarial Loss.} Regarding the landmark transformer $ T $ as a generator, we introduce the discriminator $ D_{r} $ to judge the realness of transformed landmarks, which is constrained by:
\vspace{-2mm}
\begin{equation}
	\begin{aligned}
		\mathcal{L}_{D}=\log D_{r}(L_{s,q}) + \log (1-D_{r}(\hat{L}_{s,q})),
	\end{aligned}
	\vspace{-2mm}
\end{equation}
where $ L_{s,q} $ denotes the ground truth corresponding to $\hat{L}_{s,q}$.

\begin{figure}[t] 
	\includegraphics[width=\linewidth]{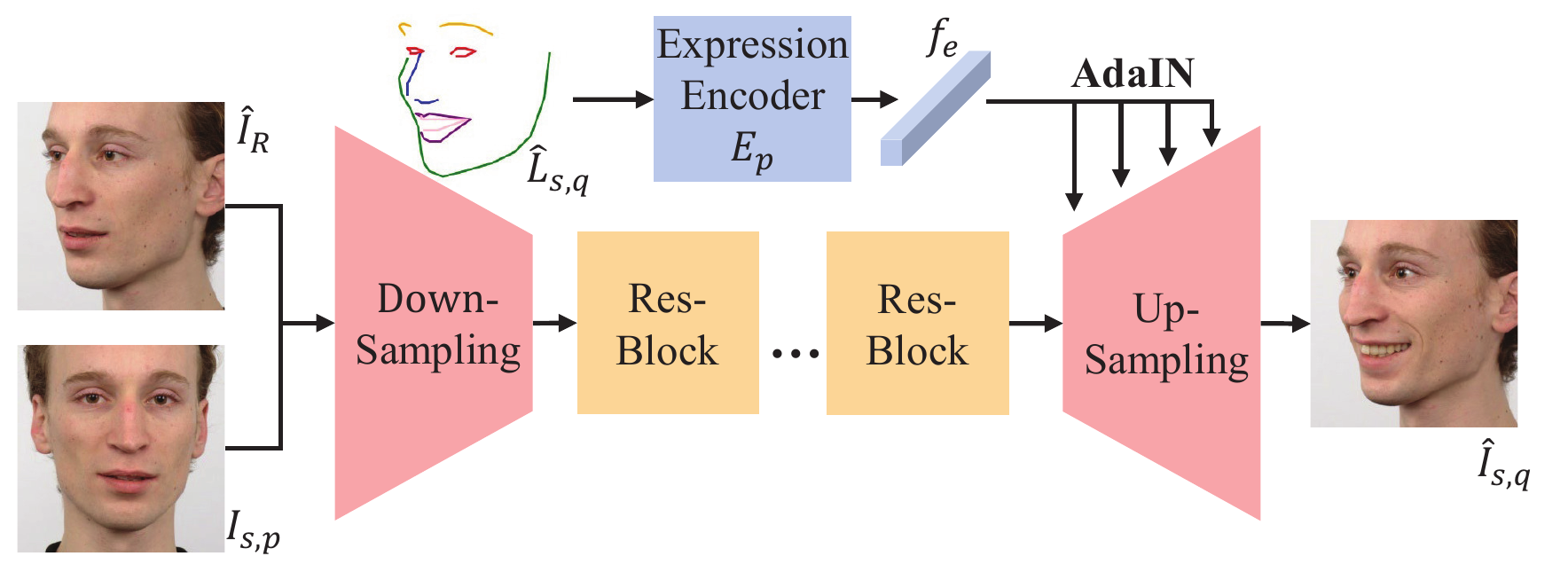} 
	\vspace{-7mm}
	\caption{\textit{\textbf{Details of expression enhancing generator $ G $.} Given the rotated face $ \hat{I}_{R} $, the source face $ \hat{I}_{s,p} $ and the expression feature extracted from $ \hat{L}_{s,q} $, G enhances $ \hat{I}_{R} $ and reenacts new expressions, generating $ \hat{I}_{s,q} $.} }
	\label{fig:expression}
	\vspace{-3mm}
\end{figure}

\vspace{-2mm}
\subsection{Face Rotation Module}
\vspace{-1mm}
\label{sec: rotation}
The face rotation module $ R $ is trained to rotate the source faces to the pose shown in $ \hat{L}_{s,q} $. It is ill-posed and difficult to directly infer a large part of the face from source face, \eg the left face of source person in Fig. \ref{fig:flowchart}. Therefore, the ultimate goal of $ R $ is not to generate photo-realistic rotated faces, but to produce a contour-consistent face image with driving pose to focus on new pose generation and reduce the burden of the entire reenacting process, \ie from  $ I_{s,p} $ to $ \hat{I}_{s,q} $.

Traditionally, when rotating the source face to a driving pose, we choose driving landmark images with driving pose that shares the same identity and expression with source face, as shown in the upper left of Fig. \ref{fig:rotator}. However, for the driving landmarks of face rotation task, we argue that only the face contour plays the most important role to decide the direction of faces while expressions and other facial parts can be relatively ignored. Hence, we propose a data augmentation method as shown in upper right of Fig. \ref{fig:rotator}. We choose landmark images with driving pose and source identity but with different expressions and facial parts. When training, we randomly choose one from those and extract the pose feature, which will be sent into R along with source face $I_{s,p}$ to generate $ \hat{I}_{R} $, as shown in the bottom half of Fig. \ref{fig:rotator}. In general, the loss terms are as follows:

\vspace{-3mm}
\begin{equation}
	\mathcal{L}_{R}= \lambda_{diff}\mathcal{L}_{diff}  +  \lambda_{GAN}\mathcal{L}_{GAN} + \lambda_{pose}\mathcal{L}_{pose},
	\vspace{-1mm}
\end{equation}
where $ \lambda $ are weights of loss terms. $ \mathcal{L}_{diff} $ minimizes the difference between rotated image $ \hat{I}_{R} $ and its ground truth $ {I}_{R} $.

\noindent\textbf{GAN Loss.} Taking the rotation module $ R $ as a generator, we apply the discriminator loss function based on LSGAN~\cite{mao2017least}:
\vspace{-2mm}
\begin{equation}
	\begin{aligned}
		\mathcal{L}_{GAN}=&  \mathbb{E}_{\boldsymbol{x} \sim p_{\text {data }}(\boldsymbol{x})}[ (D(\boldsymbol{x}))^{2}] +\\ &\mathbb{E}_{\boldsymbol{z} \sim p_{\text {data }}(\boldsymbol{z})}[ (D(R(\boldsymbol{z})) - 1)^{2} ].
	\end{aligned}
	\vspace{-2mm}
\end{equation}
where $ x $ denotes real face image data space and $ z $ denotes the input data space of the face rotation module $R$.

\noindent\textbf{Pose Prediction Loss.} To improve the accuracy of face pose synthesis, we employ another discriminator $ D_{p} $ to predict the pose of a given landmark, which is constrained by:
\vspace{-2mm}

\begin{equation}
	\begin{aligned}
		\mathcal{L}_{D_{p}}=& \mathbb{E}_{\boldsymbol{x} \sim p_{\text {data }}(\boldsymbol{x})}[( D_{p}(\boldsymbol{x})  - p)^{2}],\\
		\mathcal{L}_{pose}=& \mathbb{E}_{\boldsymbol{z} \sim p_{\text {data }}(\boldsymbol{z})}[( D_{p}(R(\boldsymbol{z}))  - p)^{2}],\\ 
	\end{aligned}	
	\vspace{-1mm}
\end{equation}
where $p$ denotes the driving pose vector. $x$ and $z$ follow the same meaning in GAN Loss. $ \mathcal{L}_{D_{p}} $ constrains $ D_{p} $ to predict the correct pose while $ \mathcal{L}_{pose} $ pushes module $ R $ to generate faces with driving pose.

\vspace{-2mm}
\subsection{Expression Enhancing Generator}
\vspace{-1mm}
\label{sec: enhancing}
As shown in Fig. \ref{fig:expression}, given the rotated source face $ \hat{I}_{R} $ and the transformed landmark $ \hat{L}_{s,q} $, the expression enhancing generator $ G $ aims to enhance detailed information  and reenact expression contained in $ \hat{L}_{s,q} $ to finally get reenacted face image $ \hat{I}_{s,q} $. An expression encoder $ E_{e} $ is utilized to encode expression information, which is later injected in residual blocks by AdaIN~\cite{huang2017arbitrary} layers. During training, the expression enhancing generator G simply changes the expression while preserving the identity and driving pose. The total loss $\mathcal{L}_{G}$ is defined as:
\vspace{-2mm}
\begin{equation}
	\mathcal{L}_{G}= \lambda_{pix}\mathcal{L}_{pix} + \lambda_{per}\mathcal{L}_{per} + \lambda_{adv}\mathcal{L}_{adv}, \label{equ:g_total_loss}
	\vspace{-2mm}
\end{equation}
where $\lambda$ are weights of various loss functions. We use $\mathcal{L}_{pix}$ as pixel-wise L1 loss to minimize the pixel difference between the rotated image $\hat{I}_{s,q}$ and the ground truth image ${I}_{s,q}$. The term $ \mathcal{L}_{adv} $ is used to improve the realness of reenacted image $ \hat{I}_{s,q}  $ in an adversarial way using traditional GAN loss.

\noindent\textbf{Perceptual Loss.} $ \mathcal{L}_{per} $ is the perceptual loss for minimizing the semantic difference of images in feature level:
\vspace{-1mm}
\begin{equation}
	\begin{aligned}
		\mathcal{L}_{per}=& \sum_{l \in \Phi}\left\|\phi_{l}(\hat{I}_{s,q}     )-\phi_{l}(I_{s,q})\right\| +\\& \sum_{l \in \Psi}\left\|\psi_{l}(\hat{I}_{s,q}     )-\psi_{l}(I_{s,q})\right\|,
	\end{aligned}
	\vspace{-1mm}
\end{equation}
where $ \Phi $ and $ \Psi $ are collections of convolution layers from the perceptual networks while $ \Phi_{l} $ and $ \Psi_{l} $ are activation outputs from $ l\text{-th} $ layers. In order to constrain that reenacted images $\hat{I}_{s,q}$ share same semantic and identity information as ${I}_{s,q}$, two perceptual networks are utilized, which are VGG-19~\cite{simonyan2014very} for image classification and  VGGFace~\cite{cao2018vggface2} for face verification.

\vspace{-2mm}
\section{Experiments}
\vspace{-1mm}
\subsection{Experimental Settings}
\vspace{-1mm}
\noindent\textbf{Datasets.} 
We use Radboud Faces Database (RaFD)~\cite{langner2010presentation} and Multi-PIE~\cite{gross2010multi} Database for experiment. RaFD has 8040 images of 67 persons displaying 8 emotional expressions, each in five different angles. Faces are cropped to 256$\times$256 and landmarks are detected by ~\cite{bulat2017far}. Faces of 55 persons are randomly selected as training set and the others test set, which means no identity overlap between two sets. Multi-PIE contains more than 750,000 images of 337 subjects, captured under 15 view points and 19 illumination conditions in four recording sessions. We use a subset of 75,000 images (2 illumination conditions) to reduce training time.

\noindent\textbf{Implementation Details.} 
Our method is implemented using PyTorch 1.5.1 on Ubuntu 18.04 with a Tesla V100 GPU. Three modules in LI-Net are trained separately using the Adam optimizer~\cite{kingma2014adam}. Landmark transformer $ T $ is trained for 2,000 epochs with batch size 128 and starting learning rate $ 1e^{-5} $. We train face rotation module $ R $ and expression enhancing generator $ G $ for 500 epochs with batch size 32. The initial learning rate is $ 2e^{-4} $ and decays by ten every 100 epochs. The $\lambda$ unifies losses on the same order of magnitude. All three modules are trained separately and detailed training schemes are described in Section \ref{sec:methodology}.

\begin{figure}[t] 
	\includegraphics[width=\linewidth]{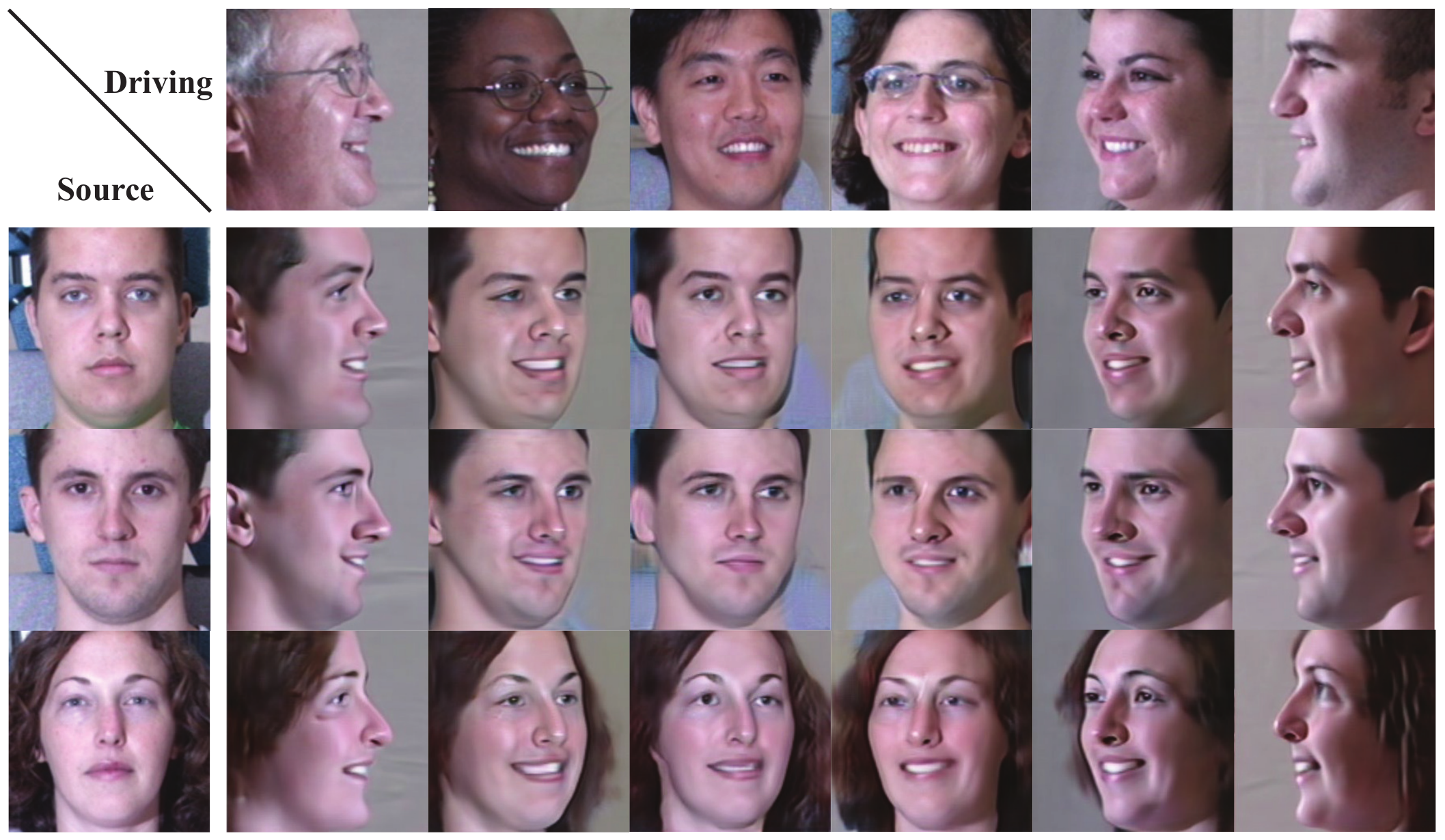} 
	\vspace{-7mm}
	\caption{\textit{\textbf{Qualitative results on Multi-PIE dataset.}}}
	\label{fig:multipie}
	\vspace{-5mm}
\end{figure}

\vspace{-2mm}
\subsection{Experimental Results}
\vspace{-1mm}
\noindent\textbf{Methods.}
We choose X2Face~\cite{wiles2018x2face}, Pix2Pix~\cite{pix2pix2017}, FOMM~\cite{siarohin2019first} and FReeNet~\cite{zhang2020freenet} as comparison methods. X2Face uses image warping to warp generated embedded face according to extracted driving expression vectors. For Pix2Pix, we follow typical image translation training scheme and concatenate source faces and driving landmark images as input. The SOTA method FOMM decouples appearance and motion information and predicts optical flow and occlusion masks to generate reenacted faces. FReeNet~\cite{zhang2020freenet} pre-processes landmarks and directly produces reenacted faces. All methods are fine-tuned using RaFD based on their pre-trained models.

\noindent\textbf{Quantitative Results.}
We compare results of face reenactment quantitatively on RaFD dataset using two metrics, structured similarity index (SSIM)~\cite{wang2004image} and Fr$\acute{\text{e}}$chet inception
distance (FID)~\cite{heusel2017gans}. Higher SSIM and lower FID mean better reenactment results. During the experiment, for each identity in test set which serves as source faces, 400 driving landmarks from test set are randomly selected to generate reenacted faces (4,800 images totally). In this way, faces of various unseen identities, diverse expressions and poses are covered. 

\begin{figure}[t] 
	\includegraphics[width=\linewidth]{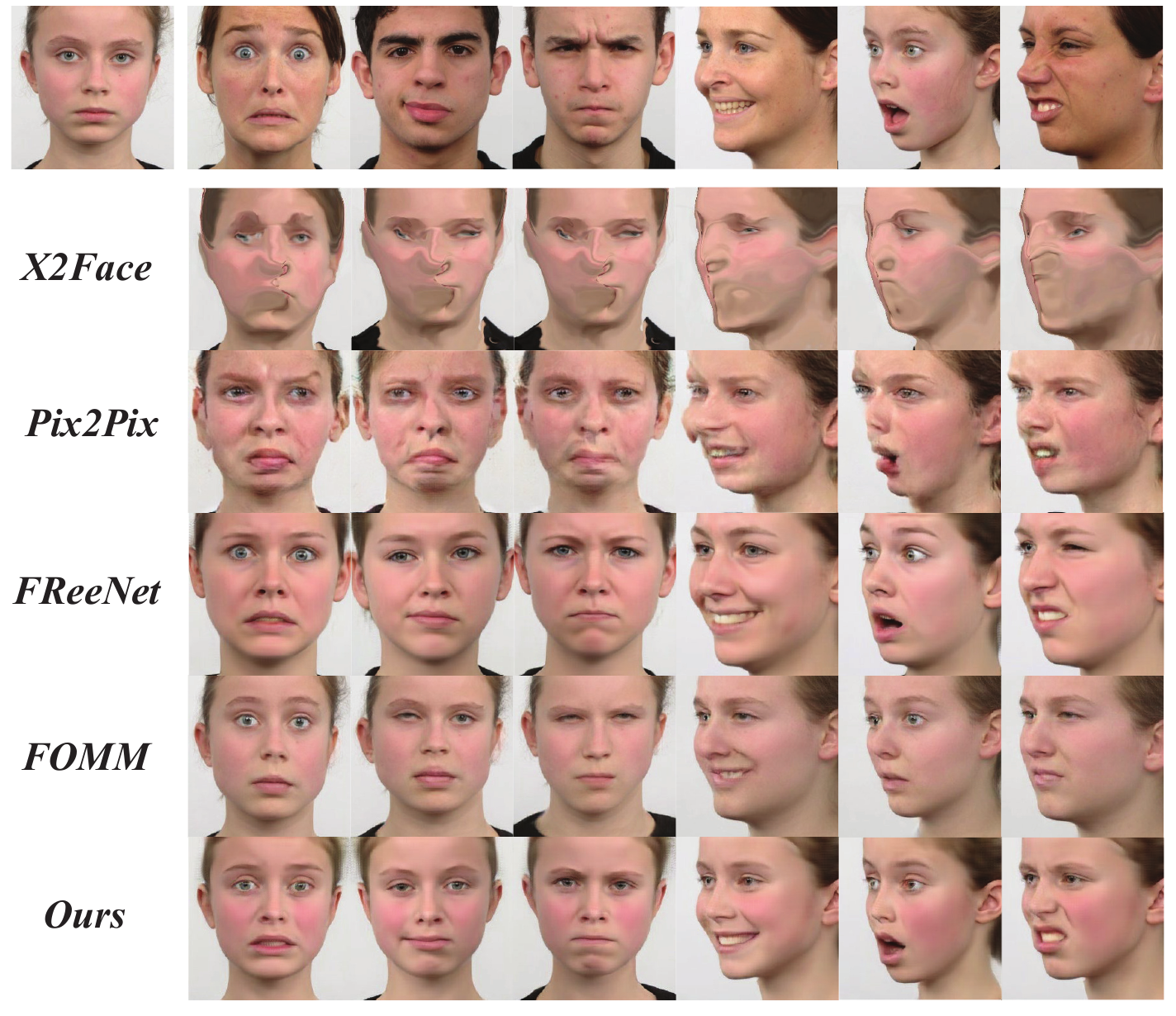} 
	\vspace{-7mm}
	\caption{\textit{\textbf{Qualitative results between LI-Net and other methods on RaFD.}} }
	\label{fig:comparison}
	\vspace{-3mm}
\end{figure}

Table \ref{tab:comparison} shows the details of quantitative evaluation, demonstrating the effectiveness of our method. Note that for FReeNet, we do not copy the original paper evaluation results, because FReeNet only adopts partial RaFD and applies test identities to training, while we use full RaFD and different split method which means no identity overlap between training and test. LI-Net gets the best SSIM score and competitive FID score compared to FOMM. It is reasonable because FID judges both variety and reality of the image. FOMM may generate various faces given that diverse landmark images are taken as guidance for one identity, unlike LI-Net. Generated face details can be seen in Fig. \ref{fig:comparison}.

\begin{table}[t]
	\caption{\textit{Metric evaluation results between other methods and LI-Net on the RaFD dateset.}} \label{tab:comparison}
	\vspace{-3mm}
	\centering
	\begin{tabular}{ccc}
		\toprule 
		\textbf {Model} & \textbf {SSIM} $\uparrow$ & \textbf {FID} $\downarrow$\\
		\hline
		X2Face~\cite{wiles2018x2face} & 0.61 & 128.43\\
		Pix2Pix~\cite{pix2pix2017} & 0.64 & 107.98\\
		FOMM~\cite{siarohin2019first} & 0.66 & \textbf{76.90}\\
		FReeNet~\cite{zhang2020freenet} & \underline{0.68} & 83.31\\
		Ours & \textbf{0.73} & \underline{80.45}\\
		\bottomrule 
	\end{tabular}
	\vspace{-5mm}
\end{table}

\noindent\textbf{Qualitative Results.}
Fig. \ref{fig:multipie} shows experimental results using Multi-PIE dataset. LI-Net can well transfer driving expressions to source persons while simultaneously maintain the correct pose and identity information. Fig. \ref{fig:comparison} shows reenacted faces between LI-Net and other methods. X2Face does not employ image realness discriminator and can not well grab face texture details simply using embedded faces and pose codes. Pix2Pix has large artifacts because the driving landmarks are used only once as the input, thus losing control to the reenacted faces when the net goes deeper. FReeNet and FOMM suffer from identity mismatch when reenacting unseen subjects, the inaccurate expression and artifacts in large pose, since they do not impose identity constraints to driving landmarks or explicitly deal with large pose situation to alleviate the inaccurate expressions or visual artifacts. Overall, LI-Net achieves the best performance in image quality while successfully preserving source identity and maintaining accurate expressions in large pose.

\begin{figure}[t] 
	\includegraphics[width=\linewidth]{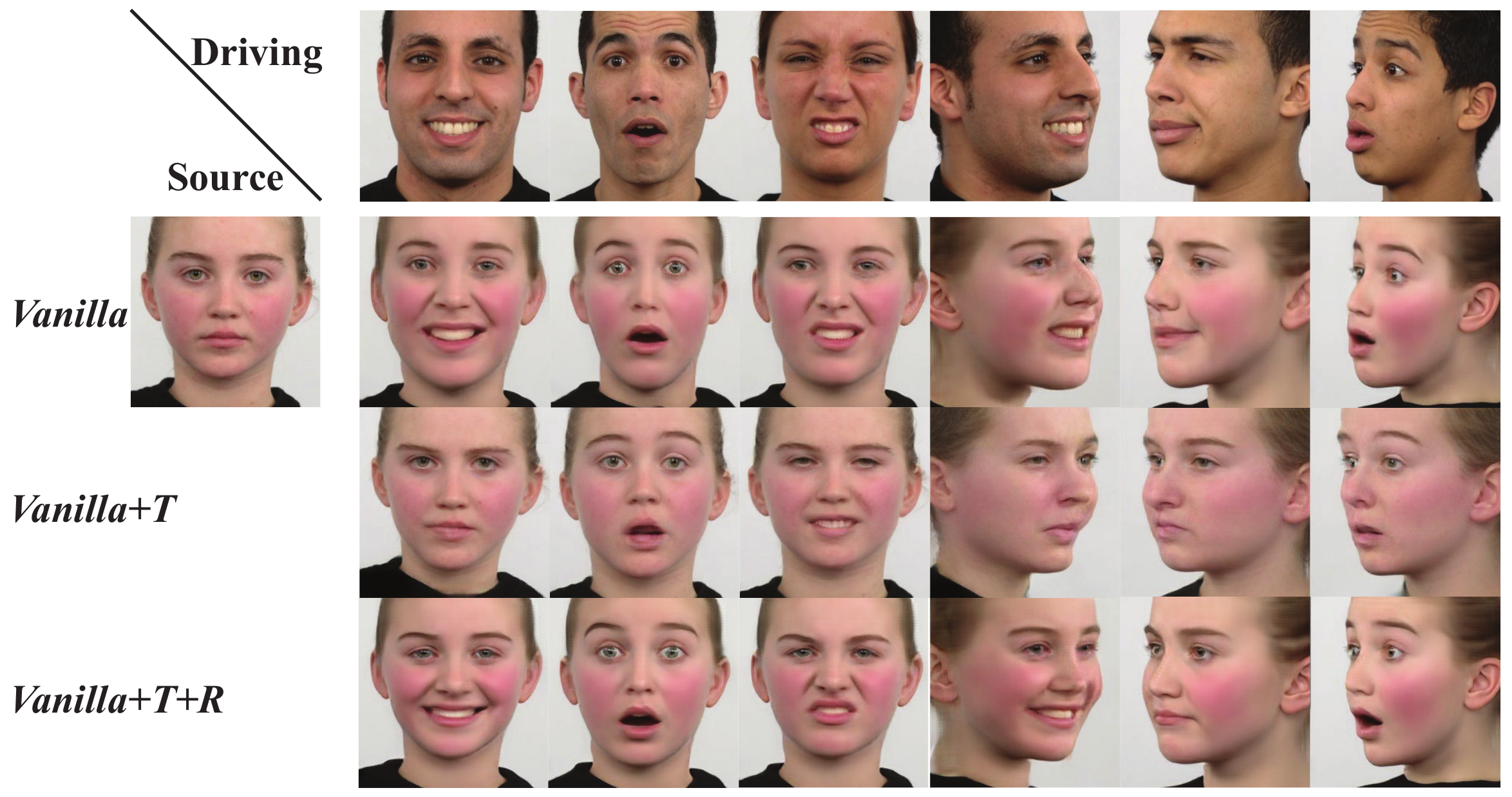} 
	\vspace{-7mm}
	\caption{\textit{\textbf{Qualitative results of ablation studies on RaFD.}} }
	\label{fig:ablation}
	\vspace{-3mm}
\end{figure}

\begin{table}[t]
	\caption{\textit{Metric evaluation results of LI-Net with different components on the RaFD dateset.}} \label{tab:ablation}
	\vspace{-3mm}
	\centering
	\begin{tabular}{ccc}
		\toprule 
		\textbf {Model} & \textbf {SSIM} $\uparrow$ & \textbf {FID} $\downarrow$\\
		\hline
		Vanilla & 0.67  & \underline{83.30} \\
		Vanilla +T & \underline{0.68} &  89.98\\
		Vanilla +T+R  & \textbf{0.73} & \textbf{80.45}\\
		\bottomrule 
	\end{tabular}
	\vspace{-5mm}
\end{table}

\noindent\textbf{Ablation Studies.}
To demonstrate the effectiveness of each module detailedly, we conduct ablation studies on three different settings using RaFD dataset: (a) \textit{vanilla}: we treat the entire process as image-to-image translation task and only utilize the expression enhancing generator $ G $. (b) \textit{vanilla+T}: based on (a), we add the landmark transformer $ T $. (c) \textit{vanilla+T+R}: our full proposed methods, \ie LI-Net.

Table \ref{tab:ablation} displays the metric evaluation results of ablation studies which shows that each component improves the SSIM score. $ T $ successfully transforms the landmarks to the source identity. $ R $ separates rotation and expression enhancing, thus improving the accuracy of facial expressions. For FID score,  $ T $ solves the identity mismatch problem but inevitably lacks the diversity in reenacted faces, which slightly gives a negative effect. When adding module $ R $, photo-realistic accurate face images are generated with more facial details, greatly improving the FID. Qualitative results are more intuitive. As shown in Fig. \ref{fig:ablation}, the face identity of vanilla results are heavily dominated by driving faces. The locations of eyes, nose, mouth between driving faces and reenacted faces are exactly the same, which is unrealistic. \textit{Vanilla+T} generates accurate source identity and maintains source face contour. However, it causes inaccurate fuzzy expressions and visual artifacts in large pose when driven by the transformed landmark. When adding R, we get faces with less artifacts, more accurate expressions and face contours in large pose which can be found between \textit{Vanilla+T} and \textit{Vanilla+T+R} results. Overall, we see a significant contribution of each component.

\vspace{-2mm}
\section{Conclusion}
\vspace{-2mm}
We present a Large-pose Identity-preserving face reenactment network called LI-Net to solve the identity mismatch problem and to generate accurate expressions with no artifacts in large-pose reenacted faces. We propose the Landmark Transformer to transform driving landmark images and adopt separate modules to generate poses and expressions individually, leading to fine-grained control over the reenacted faces. Experimental results have demonstrated that our work achieves a good performance in large-pose identity-preserving face reenactment. In future work, we plan to extend our framework to handle faces in the wild with complex backgrounds and arbitrary expressions.

\vspace{-2mm}
\bibliographystyle{IEEEtranS}
\bibliography{icme2021template}
\vspace{-2mm}

\end{document}